\documentclass[journal]{IEEEtran}

\usepackage{cite}

\ifCLASSINFOpdf
   \usepackage[pdftex]{graphicx}
\else
  
\fi

\usepackage{amsmath}
\usepackage{algorithm}  
\usepackage{algorithmic}

\usepackage{array}

\ifCLASSOPTIONcompsoc
  \usepackage[caption=false,font=normalsize,labelfont=sf,textfont=sf]{subfig}
\else
  \usepackage[caption=false,font=footnotesize]{subfig}
\fi

\usepackage{url}

\usepackage{color}
\usepackage{multirow}
\usepackage{threeparttable}
\usepackage{stmaryrd}

\begin{document}

\title{Automatic Vocabulary and Graph Verification for Accurate Loop Closure Detection}
%
%

\author{Haosong~Yue,~\IEEEmembership{Member,~IEEE,}
	Jinyu~Miao,
	Weihai~Chen,~\IEEEmembership{Member,~IEEE,}
	Wei~Wang,\\
	Fanghong~Guo,
	and Zhengguo~Li,~\IEEEmembership{Senior Member,~IEEE,}
	
	\thanks{\textit{Corresponding author: Jinyu Miao.}}
	
	\thanks{H. Yue, J. Miao and W. Chen are with the School of Automation Science and Electrical Engineering, Beihang University, Beijing 100191, China (e-mail: yuehaosong@buaa.edu.cn; jinyu.miao97@gmail.com; whchen@buaa.edu.cn)}
	\thanks{W. Wang is with the Department of Computer Science, University
		of Oxford, OX1 3QD Oxford, U.K. (e-mail: wei.wang@cs.ox.ac.uk)}
	\thanks{F. Guo is with the Department of Automation, Zhejiang University of Technology, Hangzhou, China, (e-mail: fhguo@zjut.edu.cn)}
	\thanks{Z. Li is with the Robotics Department, Institute for Infocomm Research, Singapore 138632 (e-mail: ezgli@i2r.a-star.edu.sg)}
}

\markboth{Journal of \LaTeX\ Class Files}%
{Haosong \MakeLowercase{\textit{et al.}}: Automatic Vocabulary and Graph Verification for Accurate Loop Closure Detection}


\maketitle

\begin{abstract}
Localizing pre-visited places during long-term simultaneous localization and mapping, i.e. loop closure detection (LCD), is a crucial technique to correct accumulated inconsistencies. As one of the most effective and efficient solutions, Bag-of-Words (BoW) builds a visual vocabulary to associate features and then detect loops. Most existing approaches that build vocabularies off-line determine scales of the vocabulary by trial-and-error, which often results in unreasonable feature association. Moreover, the accuracy of the algorithm usually declines due to perceptual aliasing, as the BoW-based method ignores the positions of visual features. To overcome these disadvantages, we propose a natural convergence criterion based on the comparison between the radii of nodes and the drifts of feature descriptors, which is then utilized to build the optimal vocabulary automatically. Furthermore, we present a novel topological graph verification method for validating candidate loops so that geometrical positions of the words can be involved with a negligible increase in complexity, which can significantly improve the accuracy of LCD. Experiments on various public datasets and comparisons against several state-of-the-art algorithms verify the performance of our proposed approach.
\end{abstract}


%
\IEEEpeerreviewmaketitle

\section{Introduction}
\label{intro}
Intelligent robots apply simultaneous localization and mapping (SLAM) \cite{Bailey} technique to incrementally build maps of the environments while estimating their locations. As an indispensable component, loop closure detection (LCD) algorithm is implemented to detect re-observed places and correct accumulated errors during long-term exploration \cite{orbslam, orbslam2, LidarSLAM, BathymetricSLAM}. Given the low cost of cameras and the rich information provided by images, appearance-based LCD methods \cite{Galvez-Lopez, Galvez-Lopez2, Cummins, Cummins2, SLCD, Garcia-Fidaglo, Khan, BoTW, Kazmi, Tsintotas, Garcia-Fidalgo2, Arandjelovic, Lopez-Antequera, An, Noh, Yue, Hou, Sarlin, Kenshimov, An2, Lynen, Bampis3, Milford, Siam, Siagian, Ulrich} have attracted increasing attention in the last decades. These methods retrieve candidates based on visual resemblances between the query image and referenced images in the database. Global features \cite{Siagian, Ulrich, Dalal} offer compact and efficient representations of images but are sensitive to viewpoint and illumination changes. Thus, state-of-the-art LCD proposals \cite{Galvez-Lopez, Galvez-Lopez2, Cummins, Cummins2, SLCD, Garcia-Fidaglo, Khan, BoTW, Kazmi, Tsintotas, Garcia-Fidalgo2} usually apply local features \cite{Lowe, Bay, Rublee} to obtain a more robust representation. In recent years, deep learning technology has been rapidly developed. Therefore, many LCD algorithms \cite{Arandjelovic, Lopez-Antequera, An, Noh, Yue, Hou, Sarlin, Kenshimov, An2} utilize deep Convolutional Neural Network (CNN) to extract features and have achieved improved performances. 

To minimize the computational and memory consumption, especially in large-scale environments, some time-efficient schemes \cite{Hou, Sarlin, Garcia-Fidalgo2, Milford, Siam, Galvez-Lopez, Galvez-Lopez2} have been developed. Among them, the most effective one implemented in practical SLAM systems \cite{orbslam, orbslam2} is the Bag-of-Words (BoW) framework \cite{Galvez-Lopez, Galvez-Lopez2}. In this framework, a \textit{visual vocabulary} that contains generalized features or \textit{visual words}, should be trained to associate features and represent images. The scale of the vocabulary, usually determined by the depth of vocabulary tree ($L_\omega$) and the number of child nodes in each branch ($K_\omega$), reflects the abstractive and discriminative ability of \textit{visual words}, affecting the accuracy of feature association in LCD. Traditional solutions that build vocabulary in an off-line manner usually use a trial-and-error approach. They tediously test various assumptions of $L_\omega$ and $K_\omega$, and select the combination that achieves the best performance in test sets. This strategy is quite time-consuming, and more importantly, it still cannot guarantee to get the optimal vocabulary.

\begin{figure}[t]
	\centering
	\includegraphics[width=0.47\textwidth]{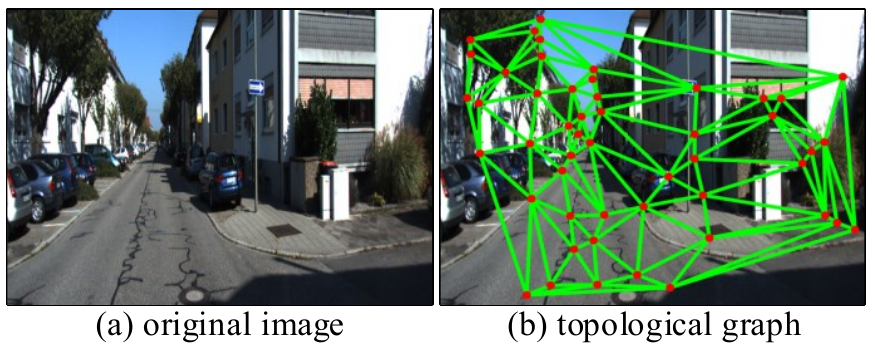}
	\caption{An example of the topological graph built by our proposed verification methods. The left (a) is the original frame, and the right (b) shows the graph. Red dots in the graph are extracted key points.}
	\label{fig:1}
\end{figure}

\begin{figure*}[t]
	\centering
	\includegraphics[width=\textwidth]{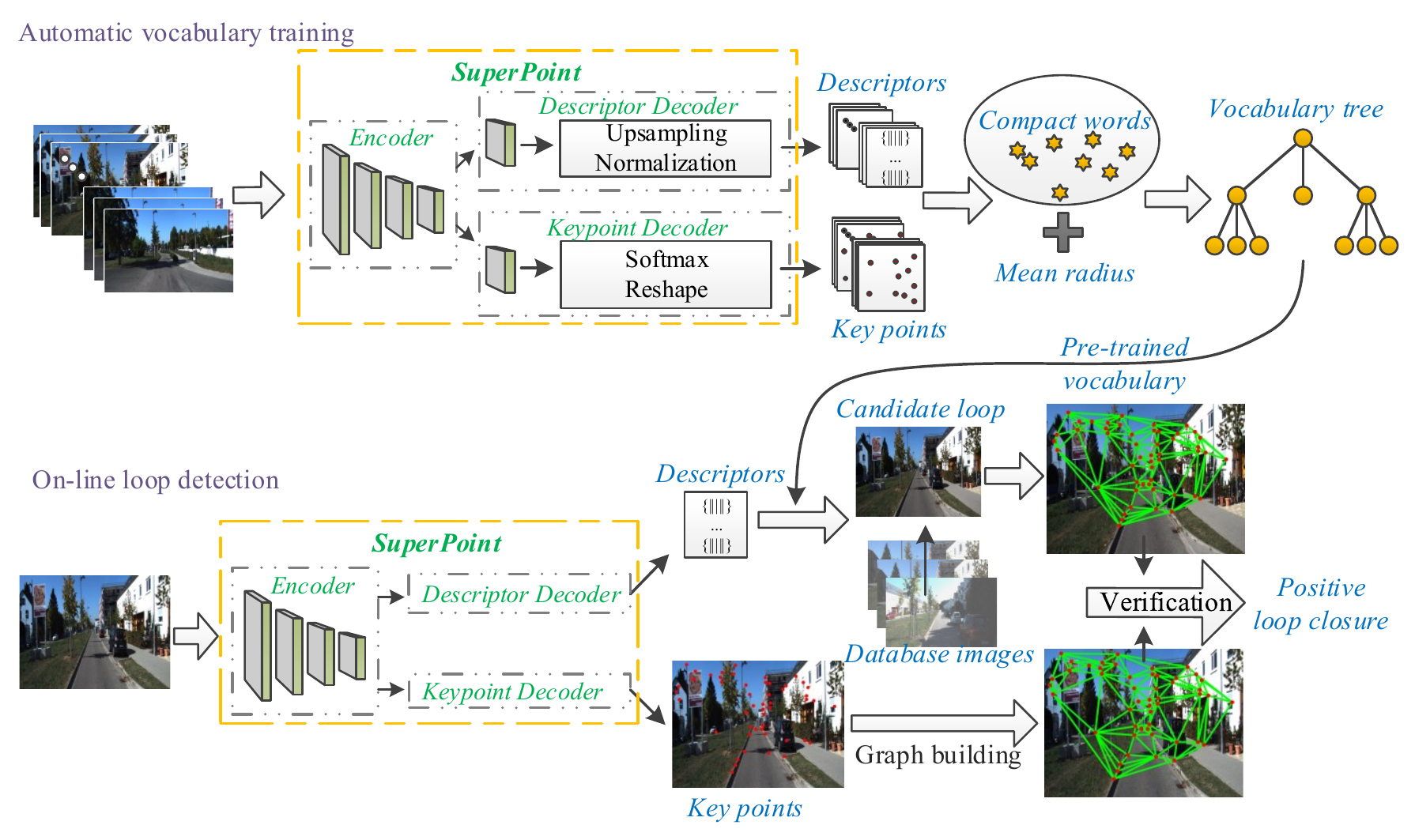}
	\caption{Overview of our proposed algorithm.}
	\label{fig:2}
\end{figure*}

BoW-based algorithms also degrade when struggling with perceptual aliasing since the spatial distribution of the words is neglected in image representation. State-of-the-art visual solutions introduce temporal consistency \cite{Khan, Milford, Siam, Garcia-Fidaglo, SLCD} or geometrical consistency \cite{Galvez-Lopez2, Garcia-Fidaglo} constraints to reject incorrect hypotheses. For example, a post-verification step based on random sample consensus (RANSAC) \cite{ransac} is always performed at the end by checking the epipolar consistency between query and loop candidates. However, temporal checks may eliminate some correct detections and decrease the LCD performance when the robot comes to the pre-visited place with a different route. Geometrical checks need sufficient correct matches between features to apply epipolar estimation and their effectiveness will be decreased when suffering from visual interference in real scenes. 

In this paper, we propose a novel BoW-based LCD algorithm, which trains a hierarchical \textit{visual vocabulary} in an automatic off-line manner and validates loops by topological graphs. The vocabulary training procedure is self-supervised and the depths of its branches can also be different. A deep CNN is utilized to extract local features from consecutive frames. Each frame is represented based on the extracted descriptors and a topological graph built by corresponding key points. The graph, as illustrated in Fig. \ref{fig:1}, can efficiently and effectively check the candidate loop closures provided by the BoW framework, significantly improving the LCD precision. Comprehensive experiments in very different datasets demonstrate that our proposed method outperforms several state-of-the-art LCD approaches with a negligible increase of the time requirement. The main contributions of this paper are summarized as follows:

\begin{itemize}
	\item An automatic vocabulary training method is proposed, which is supervised by comparison between the quantized radii of visual words and the average drift of feature descriptors. It has been experimentally demonstrated to be the optimal \textit{visual vocabulary} for BoW-based LCD.
	\item A two-step RANSAC feature matching method is proposed by considering the correspondence between sequential training images, which can produce more accurately matched features for proposed vocabulary training and estimate drifts of feature descriptors due to visual interference.
	\item A topological graph-based post-verification method is proposed to validate candidate loop closures obtained by standard BoW framework, which can be easily plugged into any appearance-based LCD algorithms.
\end{itemize}

A preliminary version of our work is presented in \cite{Yue}. In the current paper, we first design a two-step RANSAC-based feature matching method to track corresponding features in image sequences and measure the drift of feature descriptors. Then we propose the automatic vocabulary training method based on the characteristics of feature quantization, eliminating the need for a user-defined depth. We also extend the experimental evaluation of our approach and obtain more comprehensive and reliable conclusions.

The remaining of this paper is organized as follows. Section \ref{overview} provides an overall description of the proposed algorithm. Then, section \ref{voc} describes our proposed feature matching method and automatic vocabulary training strategy. A detailed description of the on-line loop detection method is presented in Section \ref{bow}. In Section \ref{test}, we present comprehensive analyses and comparative experiments on several public datasets. Finally, Section \ref{conclusion} ends with conclusions and future works.

\section{Algorithm Overview}
\label{overview}
The overview of our proposed LCD framework is illustrated in Fig. \ref{fig:2}. It consists of an off-line vocabulary training procedure (see Section \ref{voc}) and an on-line loop closure detection pipeline (see Section \ref{bow}). In the off-line procedure, a \textit{visual vocabulary} is trained automatically with the features extracted from the sequential training images. In the on-line exploration, features of the current frame are associated with \textit{visual words} and a loop closure candidate is retrieved using the standard BoW-based framework. To exclude mismatches due to perceptual aliasing, we propose a geometrical verification method based on the topological graph. A candidate is validated as a positive loop closure only if its graph is similar enough to the current image.

\label{feature}
Features matter in appearance-based LCD algorithms. In the preliminary version of our work \cite{Yue}, we demonstrated that the pre-trained SuperPoint can provide a robust image representation and improve the LCD performance. Therefore, we keep using SuperPoint to extract key points and descriptors in this paper. In our implementation, the distance of the Non-Maximum-Suppression (NMS) is set to 15 pixels to discard overlapping points. According to the heat map provided by SuperPoint, we select the top-$\mathcal{K}$ reliable features for LCD, including 2-d key points and 256-d descriptors. Moreover, the descriptors processed by the $\mathtt{StandardScale}$ method in the $\mathtt{scikit}$-$\mathtt{learn}$ library could help the convergence for vocabulary training and improve the performance. Other learned local features, such as D2-Net\cite{Dusmanu} and R2D2\cite{Revaud}, can also be utilized in our algorithm and may obtain better performance, but this is beyond the scope of this paper.

\section{Automatic Vocabulary Training}
\label{voc}
Most of the existing off-line vocabulary training procedures can be simply regarded as a clustering problem. With the user-defined parameters $K_\omega$ and $L_\omega$, these algorithms cluster image features into \textit{visual words}. Usually, a trial-and-error strategy is utilized to select the parameters that achieve good performance of LCD. However, it is time-consuming and cannot guarantee the resulting vocabulary is optimal. In this paper, an automatic vocabulary training method is proposed, which can produce a vocabulary with the optimal structure. To achieve this goal, we first present an accurate feature matching algorithm to track feature points in image sequences (see Section \ref{tracking}). Then the tracked features are ensembled into a compact database and the average drift of feature descriptors is estimated for vocabulary training (see Section \ref{database}). Finally, a natural criterion involving feature descriptor drifts is introduced to build the optimal \textit{vocabulary tree} automatically (see Section \ref{training}). Detailed descriptions are provided in the following subsections. 

\subsection{Feature Matching}
\label{tracking}
To get an accurate estimation of descriptor drifts for vocabulary training, we need to match the features corresponding to the same point in the real world, which should be associated with the same \textit{visual word}. Accurately matching features is difficult in real scenes, but the images used in LCD are generally sequential images captured by robots. Thus, the features can be tracked based on sequential spatial consistency, and accurate matches can be obtained with acceptable computational costs. Traditional tracking methods, such as LK optical flow \cite{Lucas} or KLT point tracker \cite{KLT}, degrade when appearance changes between adjacent images are large. Using conventional RANSAC method to select geometrical consistent features from two images is one of the most popular solutions, but it just eliminates wrong matches and cannot generate new matches. The trained vocabulary will be unrepresentative if there are too few image features. To obtain enough accurate matches, we propose a feature matching method based on two-step RANSAC, namely, \textit{projection} and \textit{verification}, as shown in Algorithm \ref{algo:1}.  

\begin{algorithm}[t]
	\caption{$FM(M_1,P_1,P_2,$mode$)$ $\%$feature matching}
	\label{algo:1}
	\begin{algorithmic}[1]
		\REQUIRE $M_1$:~matched~features \\
		$P_1$:~feature~point~in~$I_1$ \\
		$P_2$:~feature~point~in~$I_2$ \\
		mode:~\{projection,verficiation\} \\
		\ENSURE $M_2$:~projected/verified~matches 
		\STATE $M_2=[~]$
		\STATE $F=$calcFundamentalMat$(M_1,RANSAC)$
		\STATE $H=$calcHomographyMat$(M_1,RANSAC)$
		\IF {fail($F$)~AND~fail($H$)}{
			\STATE bad~case,~return~empty
		}
		\ELSIF {num(inlier($H$))$~>~$num(inlier($F$))}{
			\IF {mode == verification}
			\STATE $M_2$ = inliers($H$)
			\ELSIF {mode == projection}{
				\FOR {$p_1~in~P_1$}
				\STATE ${p'}_1=$HomographyProjection$(H_p,p_1)$
				\STATE $p_2=$findNearest$({p'}_1,P_2)$
				\STATE $M_2=M_2\cup[p_1,p_2]$
				\ENDFOR
			}
			\ENDIF
		}
		\ELSIF {num(inlier($F$))$~>~$num(inlier($H$))}{
			\IF {mode == verification}
			\STATE $M_2$ = inliers($F$)
			\ELSIF {mode == projection}{
				\FOR {$p_1~in~P_1$}
				\STATE $l_2=$EpipolarProjection$(F_p,p_1)$
				\STATE $p_2=$findNearest$(l_2,P_2)$
				\STATE $M_2=M_2\cup[p_1,p_2]$
				\ENDFOR
			}
			\ENDIF
		}
		\ENDIF
	\end{algorithmic}
\end{algorithm}

\begin{figure}[t]
	\centering
	\includegraphics[width=0.47\textwidth]{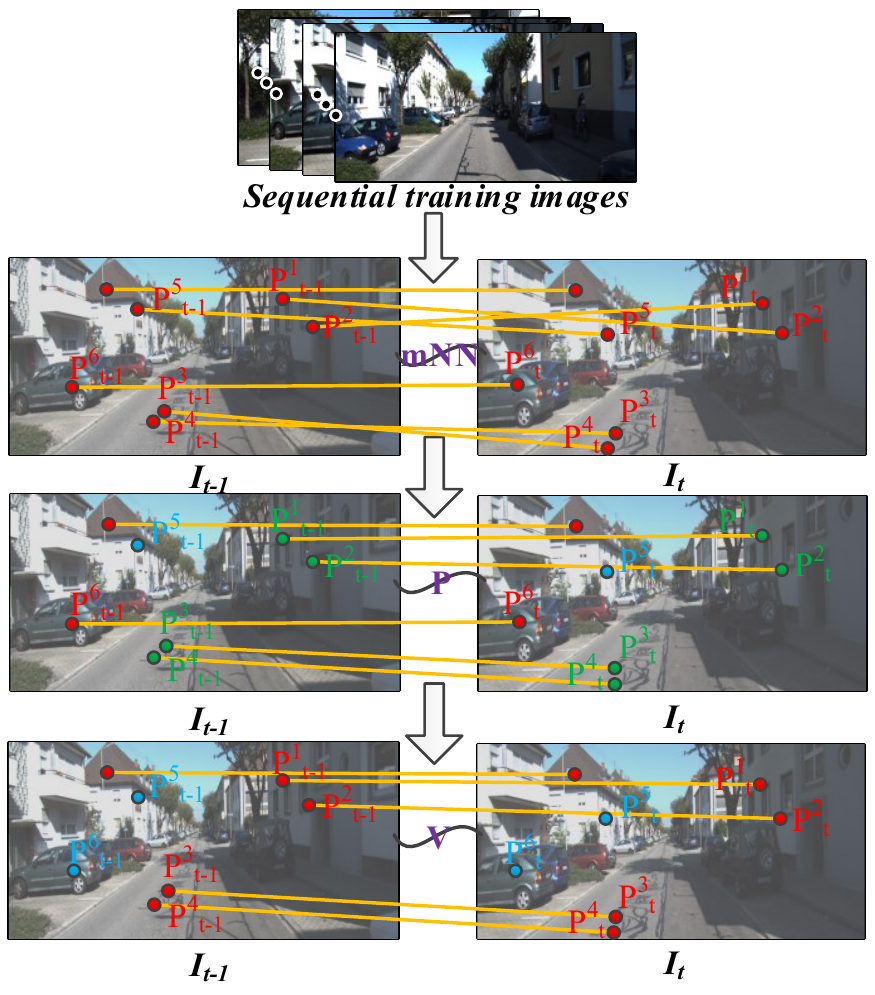}
	\caption{The main pipeline of the proposed feature matching method. The matches in the images on the top row are obtained by mutually nearest neighboring ($mNN$). The middle matches are obtained by the \textit{projection} procedure ($P$). And the matches on the bottom are obtained based on the \textit{verification} procedure ($V$). Red dots represent matched key points at each step, and blue dots represent excluded matches. The new matches generated by \textit{projection} are shown in green.}
	\label{fig:3}
\end{figure}

Let ${I}_{t}$ be the image from the training sequence at time $t$. The $\mathcal{K}$ features extracted from ${I}_{t}$ are denoted as $\mathcal{F}_{t}=\{{f}^{i}_{t}\}, i\in [1, \mathcal{K}]$, where the ${i}^{th}$ feature ${f}^{i}_{t}$ consists of a key point ${p}^{i}_{t}$ and the corresponding descriptor ${d}^{i}_{t}$. For convenience of expression, $\mathcal{{P}}_{t}=\{{p}^{i}_{t}\}$ is the set of key points. As illustrated in Fig. \ref{fig:3} , the detailed matching pipeline is described below:

\begin{enumerate}
	\item Get the matches between ${\mathcal{F}}_{t-1}$ and ${\mathcal{F}}_{t}$ using mutually nearest neighbor (mNN) matching, which is denoted as $MN=\{..., <{f}^{i}_{t-1}, {f}^{j}_{t}>, ...\}$, where $i,j$ are the indices of the features in two images, respectively.
	\item Estimate the first transform matrix between ${I}_{t-1}$ and ${I}_{t}$ by \textit{projection}, and obtain projected matches $MP$  
	\begin{equation}
	\label{equ:1}
	MP = FM(MN, \mathcal{{P}}_{t-1}, \mathcal{{P}}_{t}, {projection}).
	\end{equation}
	\item Estimate the second transform matrix by \textit{verification} using the matched features in $MP$. The inlier matches selected by the second RANSAC procedure and geometrically verified are denoted as ${M}_{V}$, which consists of the final matches of our proposed method.
	\begin{equation}
	\label{equ:2}
	MV = FM(MP,-,-,verification). 
	\end{equation}
\end{enumerate}

Both the homography ($H$) and fundamental ($F$) matrices are estimated by OpenCV functions to cover as many situations as possible, as in \cite{orbslam, orbslam2}. Furthermore, we discard unreliable matches when the numbers of inliers in calculating the $H$ and $F$ matrices are lower than 4 and 8, respectively. In \textit{projection}, the searching areas for finding nearest points around projected points (${p'}_1$) or projected epipolar lines ($l_2$) are limited to the same size of the RANSAC procedure.

In our proposed matching algorithm, RANSAC is applied first to estimate the transform matrix between adjacent images. Then the matrix is used to project the features of the former image onto the latter one to generate new matches based on geometrical consistency, which is called \textit{projection}. Features fitting such correspondence are regarded as candidate matches. As shown in Fig. \ref{fig:3}, the first RANSAC excludes the wrong matches $<{p}^{1}_{t-1},p^{2}_{t}>, <{p}^{2}_{t-1},p^{1}_{t}>, <{p}^{3}_{t-1},p^{4}_{t}>, <{p}^{4}_{t-1},p^{3}_{t}>, <{p}^{5}_{t-1},p^{5}_{t}>$. Meanwhile, new candidate matches $<{p}^{1}_{t-1},p^{1}_{t}>, <{p}^{2}_{t-1},p^{2}_{t}>, <{p}^{3}_{t-1},p^{3}_{t}>$, and $<{p}^{4}_{t-1},p^{4}_{t}>$ are generated by \textit{projection}, which cannot be obtained by using the conventional RANSAC procedure. However, $MP$ may have some incorrect results because $MN$ has mismatches due to noisy descriptors. To further validate the candidate matches, we apply RANSAC again to select the matched features that meet the second transform matrix called \textit{verification}. For instance, $<{p}^{6}_{t-1},p^{6}_{t}>$ is eliminated by the second step. More geometrically consistent matches can be obtained using our proposed method.

\subsection{Feature Aggregation and Drift Estimation}
\label{database}
By applying feature matching, a series of accurately tracked features can be obtained. Meanwhile, these features are aggregated into a \textit{compact database}, denoted as $\Psi=\{{\psi}^{i}\}$. ${\psi}^{i}$ is the ${i}^{th}$ group of the matched features from the sequential images and defined as $\psi=<\upsilon, \gamma, \chi>$, where $\upsilon$ is the center of the group, $\gamma$ is its radius, and $\chi$ is the number of contained features. $\upsilon$ can be regarded as a \textit{compact feature} and $\gamma$ denotes the largest drift of the contained feature descriptors.

Initially, the features of the first frame are extracted and added to $\Psi$ as individual groups. The initial $\upsilon$ of each $\psi$ is set as the descriptor of each feature, $\gamma$ is set to 0, and $\chi$ equals 1.

For each frame ${I}_{t}$, the features are extracted and matched with those from previous frame ${I}_{t-1}$. The matched features are updated to their corresponding groups. For example, assuming a feature ${f}_{t}=<{p}_{t}, {d}_{t}>$ from ${I}_{t}$ is matched with ${f}_{t-1}$ from frame ${I}_{t-1}$, and ${f}_{t-1}$ belongs to the ${k}^{th}$ group at time $t-1$, that is, ${\psi}^{k}_{t-1}$, then the group should be updated as 
\setlength{\arraycolsep}{0.0em}
\begin{eqnarray}
\label{equ:3}
{\upsilon}^{k}_{t} &{}={}& \frac{\upsilon^{k}_{t-1} \times {\chi}^{k}_{t-1} + {d}_{t}}{{\chi}^{k}_{t-1} + 1}, \\
\label{equ:4}
{\gamma}^{k}_{t} &{}={}& max\ ({||{d}_{i}-{\upsilon}^{k}_{t}||}_{2}),   {d}_{i}\in {\psi}^{k}_{t}, \\
\label{equ:5}
{\chi}^{k}_{t} &{}={}& {\chi}^{k}_{t-1} + 1.
\end{eqnarray}

When the database is scaled up, updating $\gamma$ will take too much time. Note that the distances between ${\upsilon}^{k}_{t}$ and the features belonging to ${\psi}^{k}_{t-1}$ change slightly after the group is updated. To save computational cost, we use the previous radius (${\gamma}^{k}_{t-1}$) to approximate such distances and avoid numerous calculations. Therefore, the formula to update the radius can be simplified as follows:
\begin{equation}
\label{equ:6}
{\gamma}^{k}_{t} = max\{{||{\upsilon}^{k}_{t} - {d}_{t}||}_{2}, {\gamma}^{k}_{t-1}\}.
\end{equation}

The unmatched features in $I_t$ are regarded as new \textit{compact features} and added into $\Psi$ as the initialization step.

After all the features from the training image sequence are incrementally ensembled into the \textit{compact database}, which is used to build the vocabulary later, the average radius $\overline{\gamma}$ is calculated as

\begin{equation}
\label{equ:7}
\overline{\gamma}=\frac{\sum_{k} \llbracket{\gamma^{k}\textgreater 0}\rrbracket \gamma^{k}}{\sum_{k} \llbracket{\gamma^{k}\textgreater 0}\rrbracket},
\end{equation}
where $\llbracket{\cdot}\rrbracket$ is the Iverson bracket, i.e. $\llbracket{}$True$\rrbracket$=1 and $\llbracket{}$False$\rrbracket=0$. The mean radius will be used as a convergence criterion for automatic vocabulary training. We approximately regard $\overline{\gamma}$ as the average drift of the feature descriptors due to appearance changes in the current scene.

\subsection{Vocabulary Building}
\label{training}

To design a natural convergence criterion that determines the scale of the \textit{vocabulary tree}, instead of user-defined parameters, we look back to the feature quantization and association in the BoW-based method and determine how the vocabulary scale affects the representative ability of \textit{visual words}. 

During vocabulary training, the entire feature space is hierarchically split into various sub-regions. The quantized ranges of nodes gradually decrease as the branch of the tree deepens, weakening the abstractive ability of nodes but strengthening their discriminative power. When detecting loops, features are associated with \textit{visual words}, that is, the leaves of the pre-trained vocabulary tree, by hierarchically retrieving the vocabulary node from the root to the bottom layer. Generally, if ${L}_{\omega}$ is extremely large, the quantized range of \textit{visual words} will be excessively small, reducing the abstractive ability of the vocabulary. In this condition, the same features from different observations may be associated with different words. In contrast, if ${L}_{\omega}$ is extremely small, different features will be associated with the same \textit{visual words}, reducing the discriminative ability.

\begin{figure}[t]
	\centering
	\includegraphics[width=0.47\textwidth]{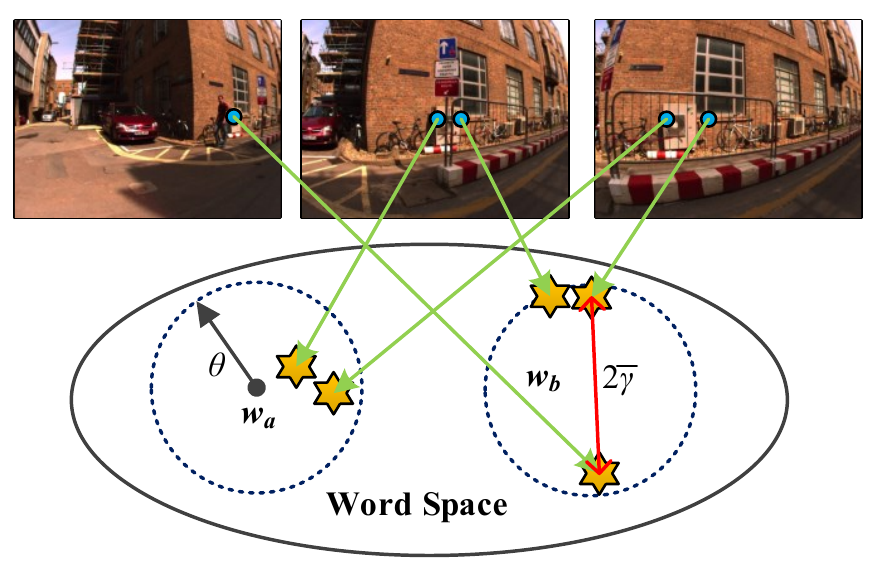}
	\caption{Schematic of optimal word quantization and feature association. The blue dots denote key points and the orange stars are the projection of their descriptors in BoW word space. Blue dashed circles $\omega_a, \omega_b$ are the resulting \textit{visual words} and $\theta$ is the quantized radius. $\theta$ should be slightly larger than $\overline{\gamma}$ so that the identical features in various visual situations can be associated with the same \textit{visual words} and meanwhile, the different features are associated with different words.}
	\label{fig:4}
\end{figure}

As shown in Fig. \ref{fig:4}, an optimal vocabulary should associate the same features in different visual conditions with the same word while distinguishing between different features. To this end, the quantized radius of visual words $\theta$ should be slightly larger than the average radius ($\overline{\gamma}$) obtained in Section \ref{database}. Therefore, we train a vocabulary tree automatically by comparing $\theta$ and $\overline{\gamma}$ of the \textit{compact database}, instead of user-defined ${L}_{\omega}$.

Assume a node in the ${d}^{th}$ layer of the vocabulary tree is $\mathcal{N}^{d}$, then its sub-node is $\mathcal{SN}^{d+1}_{p}, p\in[1, {K}_{\omega}]$ and the centers of the \textit{compact features} associated to $\mathcal{SN}^{d+1}_{p}$ are denoted as set $\{{\upsilon}^{i}_{p}\}$. During vocabulary training, we calculate the mean quantized range of sub-nodes to judge whether the branch of vocabulary tree should extend or not. For example, we apply the K-means algorithm to create ${K}_{\omega}$ sub-nodes by clustering the \textit{compact features} contained in the parent node $\mathcal{N}^{d}$ while obtaining medians ${\mathcal{MD}}_{p}, p\in[1, {K}_{\omega}]$ of each sub-node. Then, the quantized radius of each sub-node is calculated as
\begin{equation}
\label{equ:8}
{\theta}_{p} = \frac{\sum_{{\upsilon}^{i}_{p}\in{\mathcal{SN}}^{d+1}_{p}}{||{\upsilon}^{i}_{p}-{\mathcal{MD}}_{p}||}_{2}}{num({\mathcal{SN}}^{d+1}_{p})}, p\in[1, {K}_{\omega}]
\end{equation}
where $num(\mathcal{SN})$ returns the number of the \textit{compact features} contained in $\mathcal{SN}$. The L2-norm distance is used due to real-valued CNN-based features. The mean quantized radius $\overline{\theta}$ of all the sub-nodes can be calculated as:
\begin{equation}
\label{equ:9}
\overline{\theta} = \frac{1}{{K}_{\omega}}\sum_{p=1}^{{K}_{\omega}}{\theta}_{p} 
\end{equation}

If $\overline{\theta} < \overline{\gamma}$, these sub-nodes will excessively discretize features, and the extension of this branch should stop. The parent node ($\mathcal{N}^{d}$) of these sub-nodes can be defined as a \textit{visual word} (leaf node in the bottom layer). Otherwise, the branch will be extended until the criterion is met.

In our vocabulary tree training, the quantized range of the sub-nodes and the average radius of the \textit{compact database} are measured totally based on the characteristics of the features and used to control the depth of the \textit{vocabulary tree}. Thus, the overall training pipeline is self-supervised and can be automatically operated without human intervention. The training method can help to build a vocabulary with an optimal structure based on the natural convergence criterion. After obtaining the vocabulary, we associate features with the resulting \textit{visual words} and then retrieve the most similar image as loop candidate.

\section{Online Loop Detection}
\label{bow}
In this section, we describe the on-line loop detection procedure, which consists of loop candidate proposal and graph-based post-verfication.

\subsection{Candidate Proposal}
\label{candidates}
The classic BoW framework \cite{Galvez-Lopez2} is applied in our algorithm to retrieve loop closure candidates. For every query frame, we extract top-$\mathcal{K}$ key points and descriptors. The descriptors are then associated with \textit{visual words} in the trained vocabulary and the image is converted into a BoW vector ($\textbf{v}$) for matching and retrieval. The vector is a statistic histogram of the resulting \textit{visual words} that occurred in the current frame. To emphasize frequently occurred words, the term-frequency weight \textit{TF} is utilized in our work. The \textit{TF} value ${tf}^{\omega}_{t}$ of \textit{visual word} $\omega$ is calculated as follows:

\begin{equation}
\label{equ:10}
{tf}^{\omega}_{t} =\frac{{n}^{\omega}_{t}}{{N}_{t}},
\end{equation}
where ${n}^{\omega}_{t}$ is the number of occurrences of $\omega$ in image ${I}_{t}$, and ${N}_{t}$ is the total number of words that occurred in image ${I}_{t}$.

Then, the visual resemblances between images are measured by calculating the similarity between BoW vectors.  In this paper, the weighted L2-similarity is used:
\begin{equation}
\label{equ:11}
d({I}_{t}, {I}_{s}) = 1 - sqrt(1 - {({\textbf{tf}}_{t}{\textbf{v}}_{t})}^T ({\textbf{tf}}_{s}{\textbf{v}}_{s})),  s\in[1,t-\eta]
\end{equation}
where ${I}_{t}$ represents the current image at time $t$, ${I}_{s}$ represents a database image at time $s$, and ${\textbf{v}}_{t}$ and ${\textbf{v}}_{s}$ are their corresponding BoW vectors, respectively. ${\textbf{tf}}_{t}$ and ${\textbf{tf}}_{s}$ are the vectors of the \textit{TF} weights. $\eta$ is used to avoid retrieving images acquired closely. For each query image, the visited place with the highest BoW similarity is retrieved as the loop closure candidate.

\subsection{Graph Verification}
\label{verify}
As reviewed before, incorrect loop hypotheses may be produced due to perceptual aliasing because the spatial information is neglected in the image representation. The mismatches always be similar in visual contents, having similar objects in images, but the geometrical distributions of objects are different. To eliminate these mismatches, a novel topological graph-based post-verification method is proposed in this paper.

We use SuperPoint key points to build topological graphs. The key points extracted from the query image are first matched with those from the loop candidate. Then, we select the top-$T$ nearest matches to build undirected triangular graphs as too many points will increase the computational cost and decrease the robustness of verification.

\begin{figure}[t]
	\centering
	\includegraphics[width=0.47\textwidth]{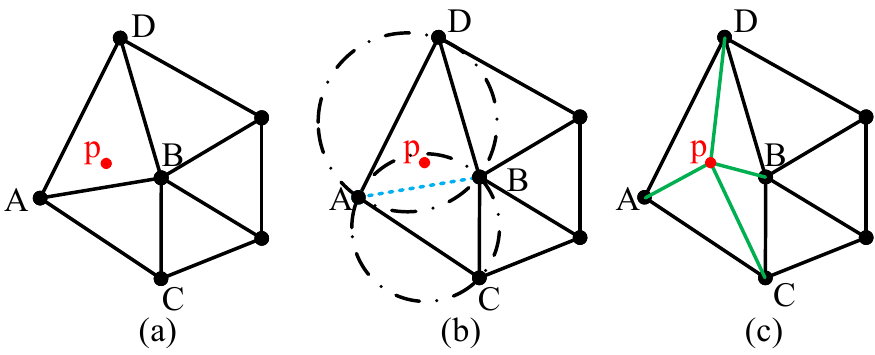}
	\caption{An example for inserting a new point into Delaunay triangular mesh.}
	\label{fig:5}
\end{figure}

In our method, Delaunay triangulation is utilized to build topological graphs, as shown in Fig. \ref{fig:1}. Several methods have been proposed to build Delaunay triangulation and the most efficient one is the Bowyer-Watson algorithm \cite{Waston}. Assuming $O=\lbrace {o}_{i}\rbrace, i\in [1,T]$ is the set of selected points ${o}_{i}$, and $E=\lbrace {e}_{j}\rbrace$ is the set of edges ${e}_{j}$, where $e_j$ represents an edge connecting adjacent points ${o}_{m}$ and ${o}_{n}, 1\leq m, n \leq T$, then the graph at the ${k}^{th}$ iteration is ${G}_{k}=\{O_k,E_k\}$. The basic procedure to build the triangulation is briefly described below:

\begin{enumerate}
	\item Assume a triangulation ${G}_{0}$ surrounding all the points.
	\item In the $k^{th}$ iteration, add a new point into ${G}_{k}$, such as ${o}_{p}$ as shown in Fig. \ref{fig:5}(a). Find all triangles whose circumcircle contains the new point, that is, $\triangle{ADB}$ and $\triangle{ABC}$, and delete public edge ${e}_{AB}$ to obtain a cavity, as shown in Fig. \ref{fig:5}(b).
	\item Connect all the vertices of the cavity with the new point ${o}_{p}$ as shown in Fig. \ref{fig:5}(c).
	\item Check whether the newly added triangles meet the requirements of Delaunay triangulation or not. If not, apply the local optimization procedure to adjust the graph and update the graph ${G}_{k}$ to ${G}_{k+1}$ .
	\item Repeat Steps 2-4 until all the points in $O$ are added to the graph. 
\end{enumerate}

The generated graph is unique regardless of the initial point, and a new feature point only affects the adjacent nodes when it is added to the graph. Therefore, this graph is insensitive to viewpoint changes. It is utilized to validate candidates obtained from the BoW framework in this paper. As we use matched features to build graphs, the vertexes in the two graphs have one-to-one matching relations. In graph matching procedure,the similarity between two topological graphs is quantified by counting their public edges (PE), that is, an edge with two matched vertexes. We randomly choose a vertex in the graph and use the breadth-first searching strategy to traverse the entire structure. The similarity between the two graphs is calculated as follows:

\begin{equation}
\label{equ:13}
\zeta({G}_{I_1}, {G}_{I_2}) = \frac{num(PE)}{num({E}_{I_1})} \cdot \frac{num(PE)}{num({E}_{I_2})},
\end{equation}
where $num(X)$ returns the number of edges in the set $X$. A larger $\zeta$ means a smaller difference between the two graphs. The loop candidate is accepted as a positive loop closure only if it has similar graph with the query image ($\zeta$ is larger than verification threshold $\zeta_t$). Otherwise, the query image has no matched loop closure or this place has not been visited.

\begin{figure*}[ht]
	\centering
	\includegraphics[width=0.97\textwidth]{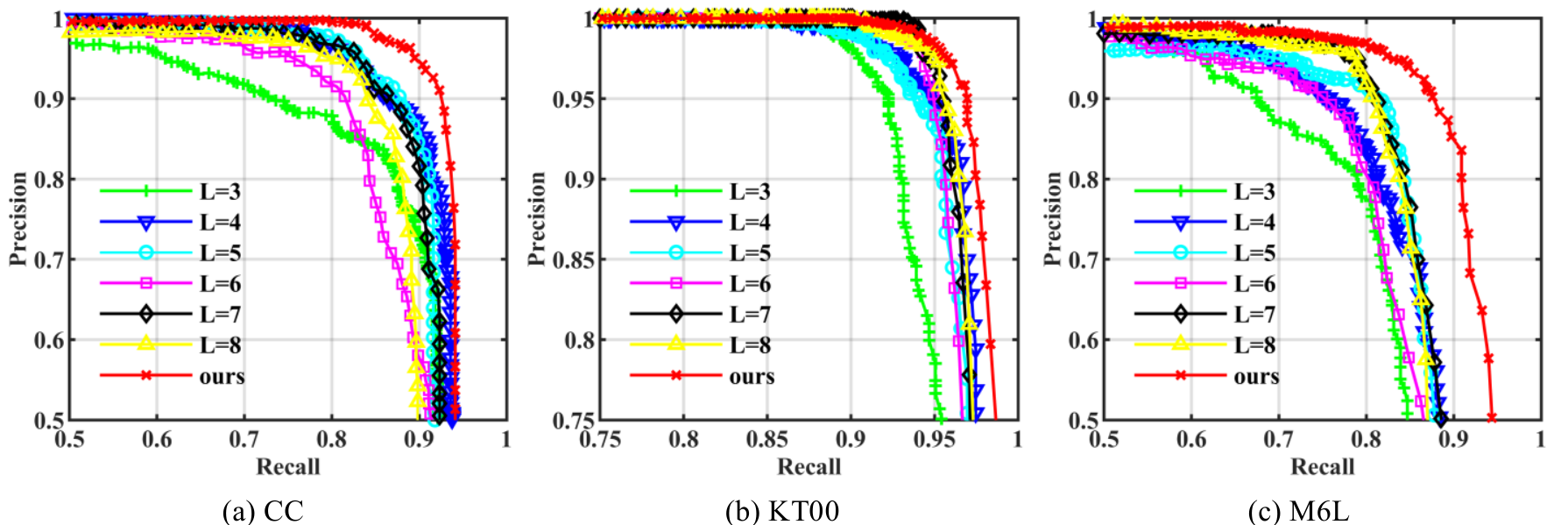}
	\caption{Comparative results with different vocabularies on various datasets. A larger area under the curve (AUC) means a better performance.}
	\label{fig:voc}
\end{figure*}

\section{Experimental Evaluation}
\label{test}
Comprehensive experiments on different datasets have been performed to evaluate our proposed algorithm. A laptop with an Intel Core i5@2.30GHz CPU and a TITAN RTX A5000 is used. In the experiments, we evaluate the performance of LCD system with commonly used precision ($P$) and recall ($R$) metrics. The publicly available SuperPoint model is utilized in our implementation. $K_\omega$, $\mathcal{K}$, and $T$ are set to 10, 500, and 50, respectively. We did not fine-tune these parameters and they may not be optimal, but the effectiveness of our proposed algorithm is still proved.

\begin{table}[b]
	\renewcommand{\arraystretch}{1.3}
	\caption{Description of datasets.}
	\label{tab:1}
	\centering
	\begin{tabular}{c c c }
		\hline
		\hline
		Dataset(abbreviation) & $\#$ Images & Description\\
		\hline
		City Centre (CC) & 1237 & Campus, slight dynamics, \\
		New College (NC) & 1073 & perceptual aliasing  \\
		\hline
		KITTI 00 (KT00)  & 4541 & \multirow{3}{*}{Urban, road, slight dynamics}  \\
		KITTI 05 (KT05)  & 2761 & \\
		KITTI 06 (KT06)  & 1101 &  \\
		\hline
		Malaga 2009 & \multirow{2}{*}{3474} & Parking lot, dynamics,  \\
		Parking 6L (M6L) & & perceptual aliasing \\
		\hline
		\hline
	\end{tabular}
\end{table}

\subsection{Datasets}
\label{method}
The proposed LCD algorithm has been evaluated on six public datasets, as listed in Table \ref{tab:1}. In City Centre \cite{Cummins} and New College \cite{Cummins}, we rectify the fish-eye images based on the calibration parameters offered by the original paper. The binocular frames are also integrated into one image for larger field-of-view. In KITTI \cite{Geiger} dataset, three typical urban sequences are used in our experiments, namely, sequences 00, 05, and 06. Only the left images in KITTI and Malaga dataset \cite{Blanco-Claraco} are utilized. All images are resized to $600 \times 800$. Approximately, 500 images are chosen at regular intervals to train a vocabulary tree in each dataset. 

In general, the ground-truth data of loops are generated by GPS logs. However, it is found that loops may occur even if the distance between two places surpasses the GPS's distance threshold. In such cases, the images still have shared appearances, and the transform matrix can be successfully estimated between two cameras. Therefore, we manually modify the ground-truth of CC and NC based on references \cite{Cummins}, and that of M6L based on \cite{BoTW}. For KITTI dataset, we directly use the public ground-truth \cite{BoTW}.


\subsection{Vocabulary Training}
\label{adpvoc}

To demonstrate that our proposed automatic training method can build a \textit{vocabulary tree} with the optimal scale, the standard BoW-based LCD algorithm\footnote[1]{DBoW3 is utilized as the basic framework in this work: \url{https://github.com/rmsalinas/DBow3}} is performed, loading different vocabularies built by our algorithm and traditional training method with various ${L}_{\omega}$. Post-verfication step is not performed. The $P$-$R$ curves are shown in Fig. \ref{fig:voc} and the numbers of visual words in each vocabulary are also provided in Table \ref{tab:voc-size}. The Recall and Precision axes begin from 0.75 in KT00, and 0.5 in M6L and CC for clear visualization.

\begin{table}[t]
	\renewcommand{\arraystretch}{1.3}
	\caption{The number of visual words in each vocabulary when $K_\omega$=10 (in thousands).}
	\label{tab:voc-size}
	\centering
	\begin{tabular}{ c | c | c c c c c c}
		\hline
		\hline
		\multirow{2}{*}{Dataset} & \multirow{2}{*}{ours ($L_\omega$)} & \multicolumn{6}{c}{traditional vocabulary $L_\omega$}\\
		& & 3 & 4 & 5 & 6 & 7 & 8\\
		\hline
		CC & 108.9 (6) & 1.0 & 10.0 & 95.6 & 253.9 & 275.7 & 277.3 \\
		KT00 & 95.8 (6) & 1.0 & 10.0 & 94.9 & 233.1 & 249.8 & 250.8 \\
		M6L & 73.0 (6) & 1.0 & 10.0 & 89.4 & 189.0 & 198.6 & 199.4 \\
		\hline
		\hline
	\end{tabular}
\end{table}

Fig. \ref{fig:voc} reveals that the vocabulary with larger scale (larger ${L}_{\omega}$ and more visual words) does not certainly perform better than the ones with smaller scales. This finding proves the motivation of the automatic training method that extremely large or small vocabularies could not quantize features well, and may decrease the performance of LCD methods. In the evaluations, the systems with ${L}_{\omega}=4$ or $5$ perform better than the others, except our method, on all the datasets.

\begin{table*}[ht]
	\renewcommand{\arraystretch}{1.3}
	\centering
	\begin{threeparttable}
		\caption{The comparative results of LCD methods without (w.o.), with graph verification, or with RANSAC-based verification (RS), and the average execution time per query of RANSAC-based and our proposed verification (in millisecond). }
		\label{tab:verification}
		
		\begin{tabular}{ c | c | c c c | c c c c c c c c c c }
			\hline
			\hline
			\multirow{2}{*}{Dataset} & \multirow{2}{*}{w.o.} &  \multicolumn{2}{c}{Inliers of RS} & \multirow{2}{*}{$\overline{t_c}$} & \multicolumn{9}{c}{Threshold of Graph Verification $\zeta_t$} & \multirow{2}{*}{$\overline{t_c}$}\\
			& & $>=$12 & $>=$20 & & 0.10 & 0.35 & 0.45 & 0.5 & 0.55 & 0.6 & 0.65 & 0.75 & 0.8 &  \\
			\hline
			CC & 0.4528 & 0.7897 & 0.8503 & 78.77 & \textcolor{red}{0.9037} & 0.8681 & 0.8378 & 0.8217 & 0.8057 & 0.7897 & 0.7736 & 0.7059 & 0.6310 & 54.32 $\downarrow$  \\
			NC & 0.4813 & 0.4813 & 0.4813 & 80.87 & 0.4813 & 0.5140 & 0.8364 & 0.8364 & \textcolor{red}{0.9463} & 0.9393 & 0.9159 & 0.8762 & 0.8551 & 53.93 $\downarrow$ \\
			KT00 & 0.8997 & 0.8997 & 0.9505 & 45.57 & 0.9505 & 0.9505 & 0.9505 & 0.9505 & \textcolor{red}{0.9569} & \textcolor{red}{0.9569} & \textcolor{red}{0.9569} & 0.9543 & 0.9492 & 31.62 $\downarrow$ \\
			KT05 & 0.9349 & 0.9349 & \textcolor{red}{0.9518} & 45.75 & 0.9349 & \textcolor{red}{0.9518} & \textcolor{red}{0.9518} & \textcolor{red}{0.9518} & \textcolor{red}{0.9518} & \textcolor{red}{0.9518} & \textcolor{red}{0.9518} & \textcolor{red}{0.9518} & 0.9446 & 31.52 $\downarrow$ \\
			KT06 & \textcolor{red}{0.9963} & \textcolor{red}{0.9963} & \textcolor{red}{0.9963} & 43.00 & \textcolor{red}{0.9963} & \textcolor{red}{0.9963} & \textcolor{red}{0.9963} & \textcolor{red}{0.9963} & \textcolor{red}{0.9963} & \textcolor{red}{0.9963} & \textcolor{red}{0.9963} & 0.9926 & 0.9890 & 31.30 $\downarrow$ \\
			M6L & 0.3934 & 0.3934 & 0.6453 & 35.52 & 0.6453 & 0.8663 & 0.8682 & 0.8798 & \textcolor{red}{0.8934} & 0.8740 & 0.8624 & 0.8295 & 0.8062 & 21.23 $\downarrow$ \\
			\hline
			\hline
		\end{tabular}
		
		\begin{tablenotes}
			\item The best performances on each dataset are emphasized in red.  $\downarrow$ means less consuming time is needed.
		\end{tablenotes}
	\end{threeparttable}
	
\end{table*}

The results also indicate that the systems loading our automatic vocabulary achieve a high recall rate and AUC even without graph verification. They perform best in all the datasets. Instead of building a regular tree with almost the same depth in each branch, we force the leaf nodes in the vocabulary to have appropriate quantized radii in our automatic training strategy. Therefore, different branches in the automatic vocabulary may have different depths, leading to a more reasonable discretization of the feature space. In Table \ref{tab:voc-size}, $L_\omega$ of our vocabularies represents the largest depth of all branches. For the vocabularies that have the same ${L}_{\omega}$, our method has the fewest words (less than half of the traditional baselines), indicating a compact structure and high retrieval efficiency. The $P$-$R$ results demonstrate that our vocabularies have the optimal scales for feature association and can retrieve more loop events.

\subsection{Graph Verification}
\label{topo}
\begin{figure}[t]
	\centering
	\includegraphics[width=0.47\textwidth]{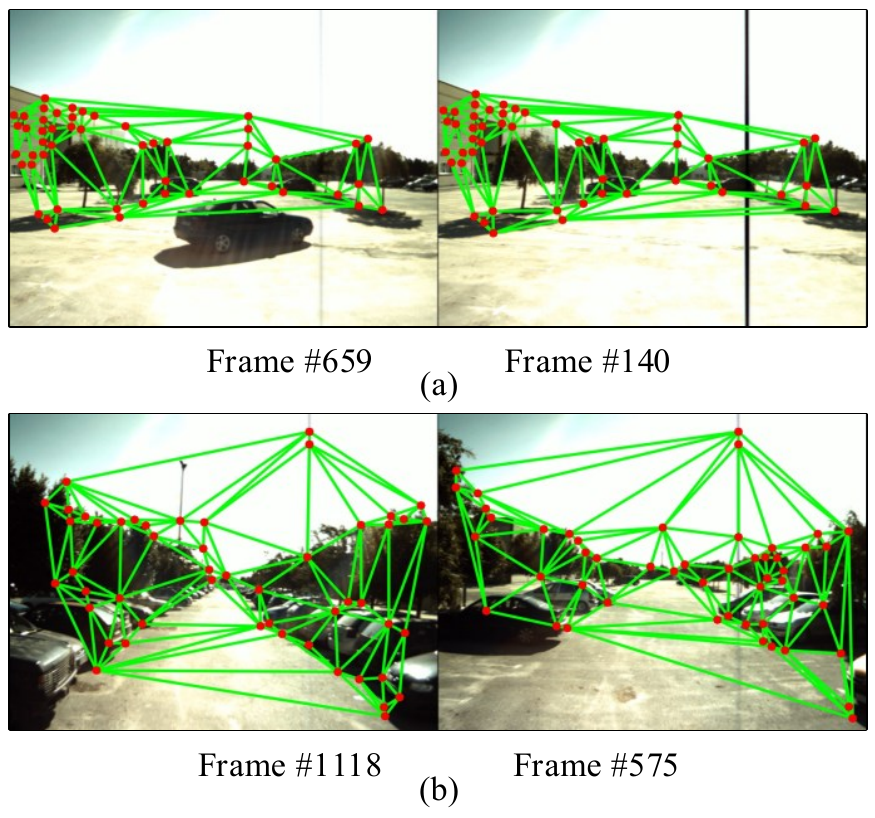}
	\caption{Topological graphs of two loop closure candidate pairs in the M6L dataset. The similarity scores between two graphs are (a) 0.9498 and (b) 0.0840, respectively.}
	\label{fig:topo}
\end{figure}

Our proposed verification method eliminates mismatches due to perceptual aliasing, by comparing the topological similarity between graphs of the query and the loop candidate. Only matched features are used to build the graphs so that the features located in dynamic regions will be discarded. As shown in Fig. \ref{fig:topo}(a), true loop closures with dynamics have highly similar graphs. In Fig. \ref{fig:topo}(b), different places with many identical \textit{visual words} will be misjudged as loop closure by BoW-based method but the similarity between their graphs is quite low so that they can be effectively excluded by our method.

To quantitatively analyze our proposed graph verification, LCD algorithm without verification (w.o.), with our proposed graph verification, and with conventional RANSAC-based verification (RS) are performed. Additionally, we also regulate $\zeta_t$ to discuss the generalization ability of the graph verification. The RANSAC-based method uses RANSAC to estimate a fundamental matrix and discard loop candidates if the number of inliers is less than 12 \cite{Tsintotas, Galvez-Lopez2} or 20 \cite{An}. Since a wrong loop closure may cause disastrous failure in building a consistent map \cite{BathymetricSLAM}, the maximal Recall when Precision achieves 100\% ($R_{max}$@$1.0$) is listed here, as in Table \ref{tab:verification}.

It can be seen from Table \ref{tab:verification} that our proposed graph verification excludes most wrong matches and significantly increase the precision of detection in all datasets. Similar objects frequently occur in CC, NC, and M6L, so the verification procedure is very necessary and relatively sensitive to the threshold. In KT05 and KT06, the improvements are slight due to fewer repetitive patterns in these scenes. There is not much room for our graph verification to present its effectiveness and it can hold its performance irrespective of the threshold. We find that 0.55 can be a general threshold among various datasets. It only fails in CC, but the performance is still comparable. The proposed verification method show satisfactory generalization ability when setting $\zeta_t=0.55$.

Compared with the RANSAC-based algorithm, our method outperforms it in most of the datasets. The RANSAC-based method focuses on all the features and it will be easily affected by the noisy absolution position of the matched feature point. Our topological graph focuses on the relative distribution of matched features so its robustness is enhanced in consequence. Moreover, our proposed verification costs less computational time ($\overline{t_c}$) than RANSAC, as listed in Table \ref{tab:verification}, as fewer matched features are needed.


\subsection{Execution Time}
\label{time}

\begin{figure}[t]
	\centering
	\includegraphics[width=0.47\textwidth]{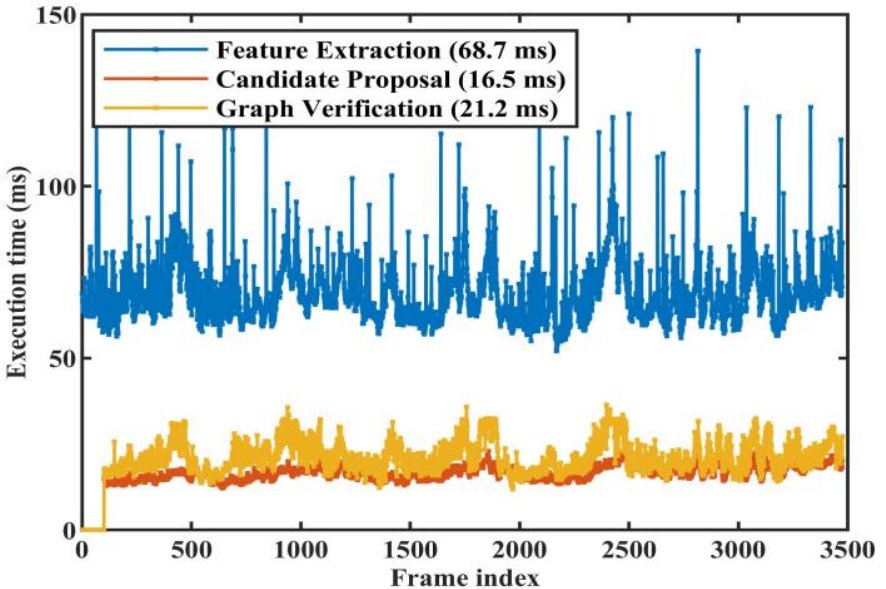}
	\caption{Execution time of each step in M6L. We begin to detect loops after $\eta=$100 images being added into database in M6L.}
	\label{fig:time}
\end{figure}

Computational efficiency of SLAM systems is quite important. To analyze the execution efficiency of our proposed LCD algorithm, it is tested on four different datasets and the mean computational times of its three steps are recorded, namely, feature extraction, candidate proposal, and graph verification, as presented in Fig. \ref{fig:time} and Table \ref{tab:time}.

\begin{table}[t]
	\renewcommand{\arraystretch}{1.3}
	\centering
	\begin{threeparttable}
		\caption{Mean Execution time of different steps per query (ms).}
		\label{tab:time}
		\begin{tabular}{ c | c c c c c}
			\hline
			\hline
			& CC &  KT00 & KT05 & M6L \\
			\hline
			Feature extraction & 146.8 & 81.2 & 80.6 & 68.7  \\
			Candidate proposal & 35.8 & 21.4 & 19.9 &  16.5  \\
			Graph verification & 54.3 & 31.6 & 31.5 & 21.2  \\
			\hline
			Total & 236.9 & \textcolor{red}{134.2} & \textcolor{red}{132.0} & \textcolor{red}{106.4} \\
			\hline
			iBoW-LCD 2018 \cite{Garcia-Fidaglo} & 1418.2 & 1695.2 & 1616.0 & 1850.1 \\
			HTMap 2017 \cite{Garcia-Fidalgo2} & 245.3 & 338.7 & 290.7 & 411.1 \\
			Tsintotas \textit{et al.} 2019 \cite{BoTW} & 490.7 & 1329.9 & 1268.3 & 1075.2  \\
			Tsintotas \textit{et al.} 2018 \cite{Tsintotas} & \textcolor{red}{232.8} & 383.4 & 297.8 & 416.0  \\
			preliminary work 2019 \cite{Yue} & 241.4 & 138.1 & 135.3 & 108.8 \\
			\hline
			\hline
		\end{tabular}
	\end{threeparttable}
\end{table}

The table indicates that the CNN-based feature extraction costs a significant amount of computational time in our proposed algorithm, which can be accelerated by a better GPU or replaced by other light features. By applying efficient BoW framework, our approach could update the database and then retrieve loop candidates efficiently in approximately 20 ms. The execution times in CC and NC are approximately 2 times higher than those of the others because we use binocular images and 1000 features per query (two images). The proposed topological graph-based verification is also proven to be efficient. Moreover, the computation time does not increase as the database scales up as shown in Fig. \ref{fig:time}.

Comparative results with existing methods are also provided in Table \ref{tab:time}. The results of compared baselines are obtained by our implementation using open-sourced codes with default configurations. Although our method utilizes real-valued CNN-based features, which consumes a significant computational time, the totally required time remains less than the baselines \cite{Garcia-Fidaglo, Tsintotas, BoTW, Garcia-Fidalgo2} in most datasets. The execution time of the proposed algorithm is slightly less than that of our preliminary work \cite{Yue} mainly due to the compact vocabulary, which proves that our automatic vocabulary can reduce computational consumption.

\subsection{Performance of Overall Algorithm}
\label{compare}
To comprehensively evaluate our proposed LCD algorithm, it is compared with many state-of-the-arts appearance-based LCD methods on six public datasets. For quantitative comparison, $R_{max}$@$1.0$ is recorded in Table \ref{tab:compare}. In this evaluation, we first load the individual vocabularies automatically trained on each dataset and fine-tune the threshold of graph verification for each dataset, which indicates an upper-bound performance of our proposed approach. The fine-tuned configurations are also shown in Table \ref{tab:compare}.

\begin{table*}[t]
	\renewcommand{\arraystretch}{1.3}
	\centering
	\begin{threeparttable}
		\caption{Comparative results with state-of-the-art appearance-based LCD algorithms. }
		\label{tab:compare}
		
		\begin{tabular}{c | c c c c c c c c | c c}
			\hline
			\hline
			\multirow{2}{*}{Test set} & DLoopDetector & HTMap & Tsintotas & iBoW-LCD & Kazmi\tnote{\dag} & FILD\tnote{\P} & Yue & Chen\tnote{\dag}  & \multicolumn{2}{c}{proposed} \\
			& \cite{Galvez-Lopez2} 2012 & \cite{Garcia-Fidalgo2} 2017 & \cite{Tsintotas} 2018 & \cite{Garcia-Fidaglo} 2018 & \cite{Kazmi} 2019 & \cite{An} 2019 & \cite{Yue} 2019 & \cite{SLCD} 2021 & general & fine-tuned($\zeta_t$) \\
			\hline
			CC & 0.2389 & 0.6292 & 0.6132 & 0.6747 & 0.7558 & 0.6648 & \textcolor{green}{0.8984} & 0.4088 & \textcolor{blue}{0.7932} & \textcolor{red}{0.9127} (0.08) \\
			NC & 0.3551 & 0.7360\tnote{$\star$} & 0.1108 & 0.6747 & 0.5109 & n.a. & \textcolor{green}{0.9299} & 0.7529 & \textcolor{blue}{0.8481} & \textcolor{red}{0.9463} (0.55) \\
			KT00 & 0.7243 & 0.9024\tnote{$\star$} & 0.9318 & 0.5774 & 0.9039 & 0.9123 & \textcolor{blue}{0.9416} & \textcolor{red}{0.9753} & 0.9353 & \textcolor{green}{0.9569} (0.55) \\
			KT05 & 0.5197 & 0.7588\tnote{$\star$} & \textcolor{blue}{0.9420} & 0.5307\tnote{\S} & 0.8141 & 0.8515 & \textcolor{green}{0.9494} & 0.8972 & \textcolor{red}{0.9518} & \textcolor{red}{0.9518} (0.55) \\
			KT06 & 0.8971 & \textcolor{green}{0.9926} & 0.8603 & 0.9228 & 0.9739 & 0.9338 & \textcolor{red}{0.9963} & 0.9313 & \textcolor{blue}{0.9890} & \textcolor{red}{0.9963} (0.55) \\
			M6L & \textcolor{blue}{0.7475}\tnote{$\star$} & n.a. & 0.1321 & 0.0178 & 0.5098 & 0.5609 & \textcolor{green}{0.8605} & n.a. & \textcolor{green}{0.8605} & \textcolor{red}{0.8934} (0.55) \\
			\hline
			\hline
			
		\end{tabular}
		
		\begin{tablenotes}
			\item The \textcolor{red}{best}, \textcolor{green}{second-}, and \textcolor{blue}{third-best} performances are emphasized in red, green, and blue, respectively. \tnote{\dag}We directly report the results in \cite{Kazmi} and \cite{SLCD}, since the official implementation is not open-sourced. \tnote{\P}The results are obtained from the latest publication of the same authors \cite{An2} since the authors do not provide trained weights. \tnote{$\star$}In our implementations, the method cannot achieve 100\% precision in the dataset, so we report the result from its original paper \cite{Galvez-Lopez2, Garcia-Fidalgo2}. \tnote{\S}And we report the result in KT05 from \cite{Kazmi} for the same reason. 
		\end{tablenotes}
	\end{threeparttable}
\end{table*}

\begin{figure}[t]
	\centering
	\includegraphics[width=0.47\textwidth]{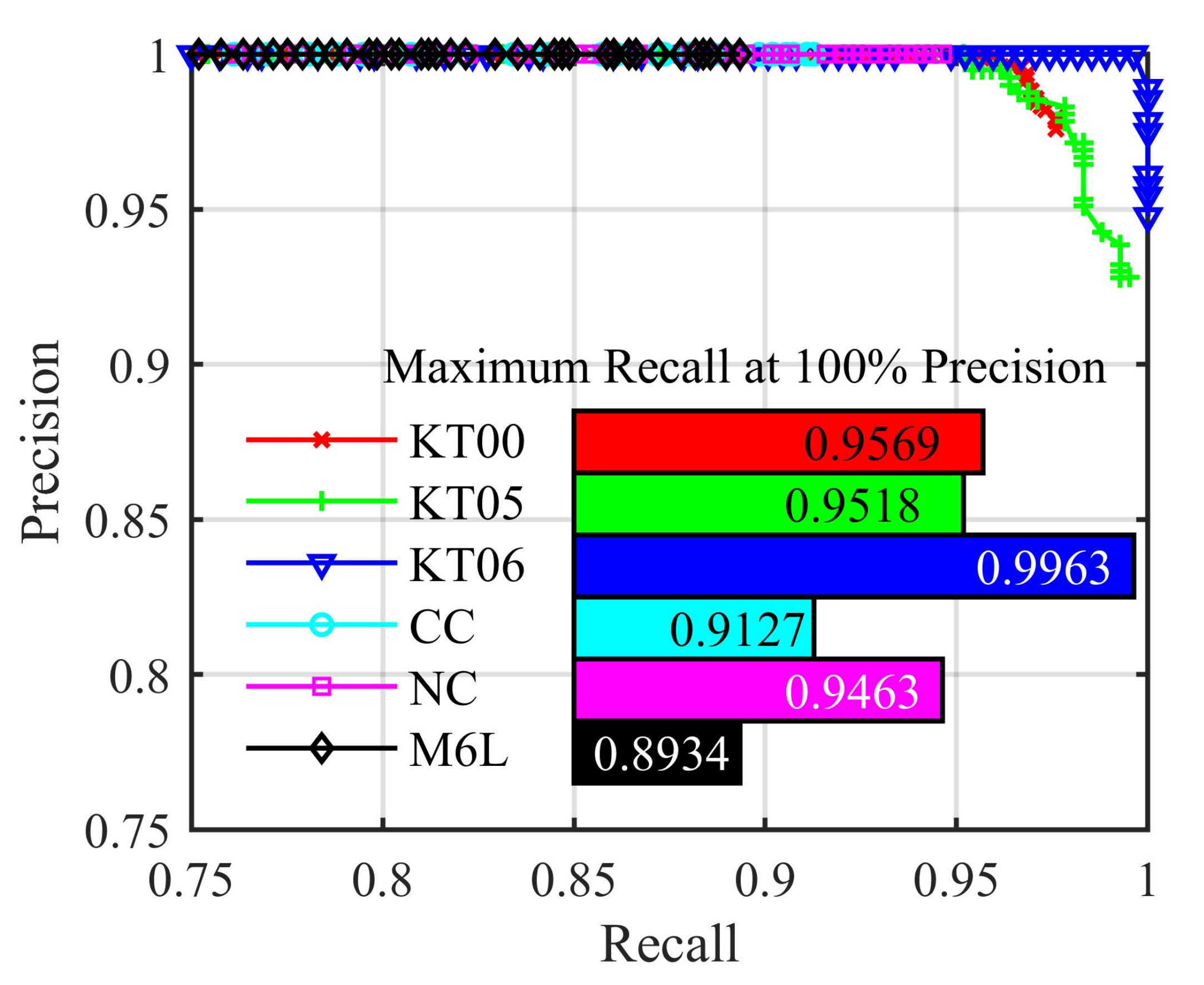}
	\caption{Fine-tuned performance achieved by our proposed algorithm in various datasets.}
	\label{fig:pr}
\end{figure}

The table shows that our proposed algorithm with topological graph verification exhibits the best or second-best performance in all datasets. The algorithm can detect more than $90\%$ loop closure events with $100\%$ precision on most of the datasets, as shown in Fig. \ref{fig:pr}, which indicates the state-of-the-art detection performance in appearance-based LCD algorithms to the best of our knowledge. 

Our system is developed based on the DBoW3 framework, which is technically similar to DLoopDetector \cite{Galvez-Lopez2}. The experimental results show that significant improvements have been achieved on all datasets since we use CNN-based features, automatic vocabularies, and graph verification. The proposed method significantly outperforms FILD \cite{An}, which is a baseline that also uses CNN-based features. In our evaluation, we also trained the individual vocabulary and tuned the threshold of graph verification in our preliminary work \cite{Yue} to obtain better performance. The proposed method exhibits better performances in all datasets than our preliminary work mainly due to the automatic vocabulary, which concludes the effectiveness of the proposed training method. 

Compared with not only the BoW-based methods building vocabulary in off-line manner \cite{Galvez-Lopez2, SLCD, Yue} but also in on-line/incremental manner \cite{Garcia-Fidaglo} , our proposed method can detect more loops accurately. Besides, our method achieves better performance than methods based on other incremental schemes \cite{Garcia-Fidalgo2, An, Kazmi, Tsintotas}, indicating that the BoW-based method with optimal vocabulary and effective post-verification can still be the best choice for practical SLAM.

Furthermore, to evaluate the generalization of the whole system, we adopt threshold $\zeta_t=0.55$ and KT05 vocabulary as a general configuration after comparing the performance of each automatic vocabulary. This vocabulary works well on all datasets and has a relatively compact structure (95.9K words). As shown in Table \ref{tab:compare}, the method with general setups also achieves a comparable performance and outperforms the methods \cite{Galvez-Lopez2, SLCD} in most of the datasets which also load a general vocabulary. Thus, our proposed LCD method exhibits a satisfactory generalization ability, which are beneficial to practical implementation in SLAM system.

\section{Conclusion}
\label{conclusion}
In this paper, a novel LCD algorithm based on automatic vocabulary and topological graph verification is proposed. SuperPoint features from image sequences are matched by a two-step RANSAC method to build a \textit{compact database} so that the average drift of the feature descriptors can be approximately estimated. Comparison between the drift and quantized range of nodes is utilized as the convergence criterion to train a \textit{vocabulary tree} automatically. To overcome the perceptual aliasing problem, a topological graph-based verification method is proposed to eliminate wrong matches detected by the standard BoW framework.

Comprehensive experiments have been performed to evaluate our proposed method. Our automatic vocabulary is proven to have an optimal and compact structure, which outperforms traditional vocabularies trained with user-defined scales. The results also demonstrate that our graph verification method can significantly improve LCD performance and perform better than conventional RANSAC-based verification. Comparative experiments on several datasets show that our proposed method achieves the highest recall at 100$\%$ precision while detecting loops efficiently.

The proposed automatic vocabulary training and graph verification can be easily plugged into any BoW-based LCD frameworks as independent components, which will further improve the performance of LCD algorithms. In the future, we would like to test our proposed LCD method in a practical SLAM system to increase the accuracy of localization and mapping.

\section*{Acknowledgment}
This research was funded by National Natural Science Foundation of China (No.61603020, No.61620106012), the Fundamental Research Funds for the Central Universities (No.YWF-21-BJ-J-923), and the Foundation of Strengthening Program Technology Fund Projects (No. 2019-JCJQ-JJ-268).

\ifCLASSOPTIONcaptionsoff
  \newpage
\fi



\bibliographystyle{IEEEtran}
\bibliography{IEEEexample.bib}

%
%
%

%

\begin{IEEEbiography}[{\includegraphics[width=1in,height=1.25in,clip,keepaspectratio]{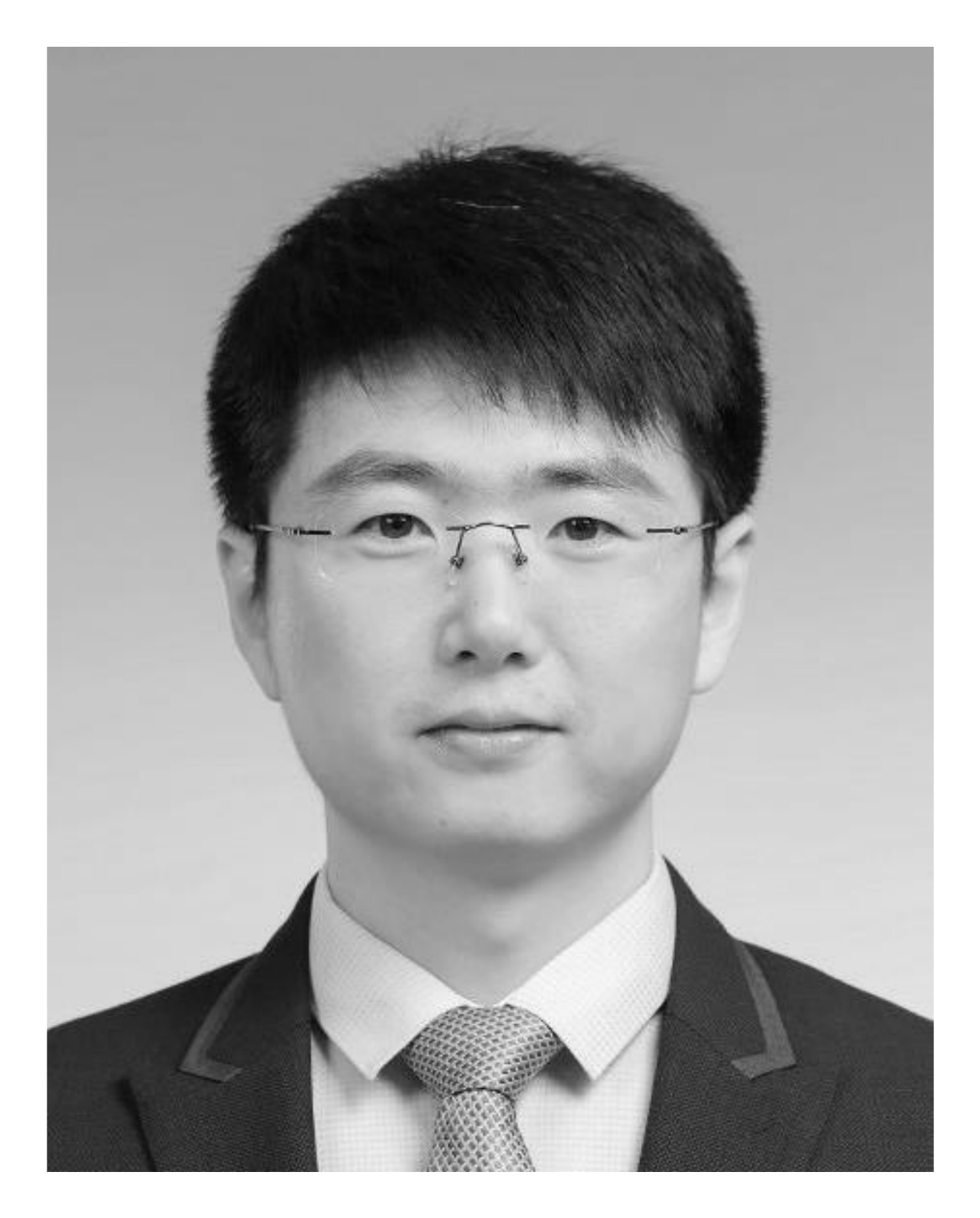}}]{Haosong Yue}(M'19)
	received the Ph.D. degree from Beihang University, Beijing, China, in 2015.
	He is currently an Assistant Professor with the School of Automation Science and Electrical Engineering, Beihang University. He	has authored more than 30 journal or conference papers. He served as a SLAM Session Co-Chair of IEEE/RSJ IROS 2019. His research interests include robot vision and control, simultaneous localization and mapping, and 3D reconstruction.
\end{IEEEbiography}
\vspace{-40pt}
\begin{IEEEbiography}[{\includegraphics[width=1in,height=1.25in,clip,keepaspectratio]{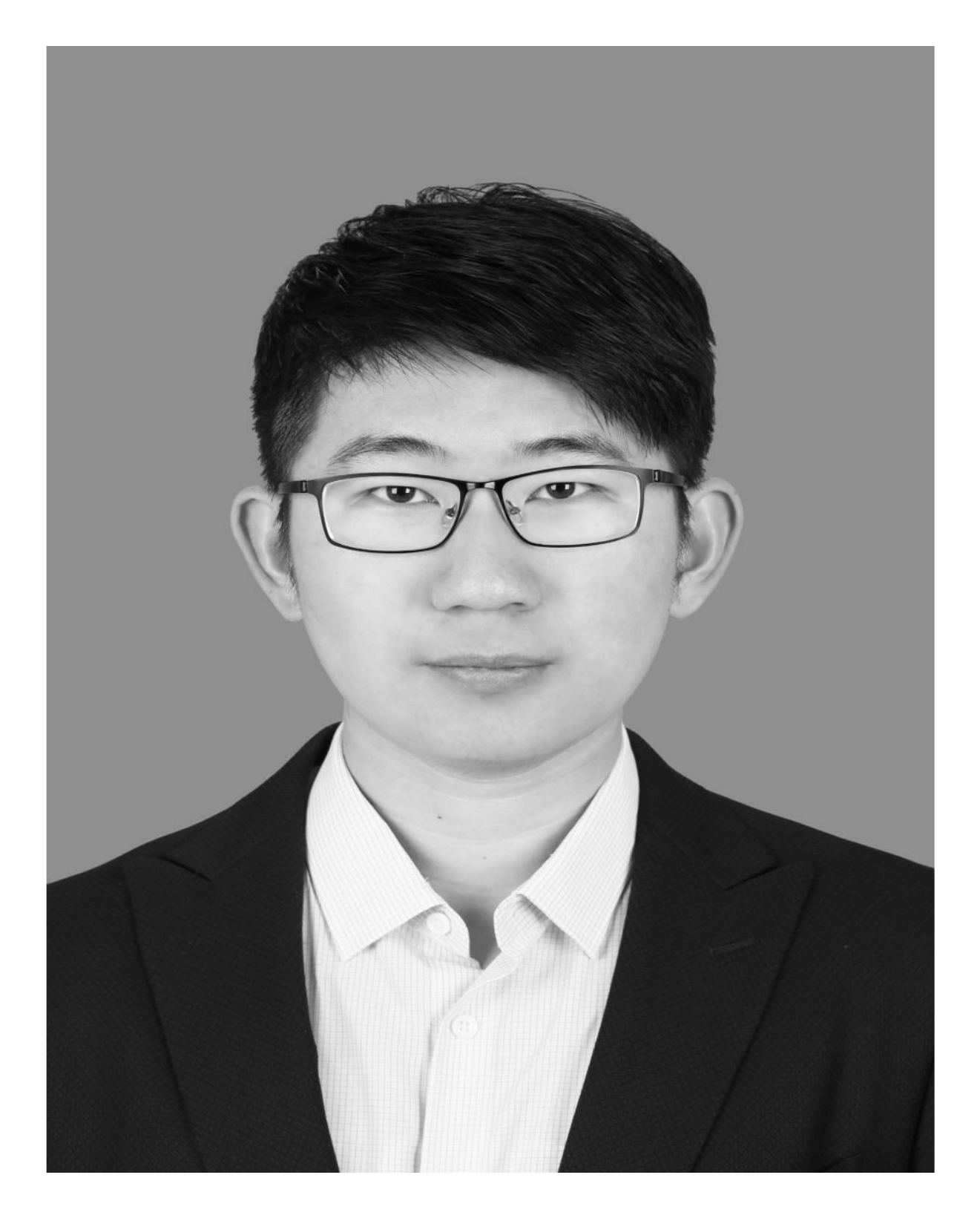}}]{Jinyu Miao}
	received the B.Sci. degree in control science and engineering from Beihang University, Beijing, China, in 2019. 
	He is currently a master in the School of Automation Science and Electrical Engineering, Beihang University, Beijing, China. His research interests include place recognition, deep learning, and simultaneous localization and mapping.
\end{IEEEbiography}
\vspace{-40pt}
\begin{IEEEbiography}[{\includegraphics[width=1in,height=1.25in,clip,keepaspectratio]{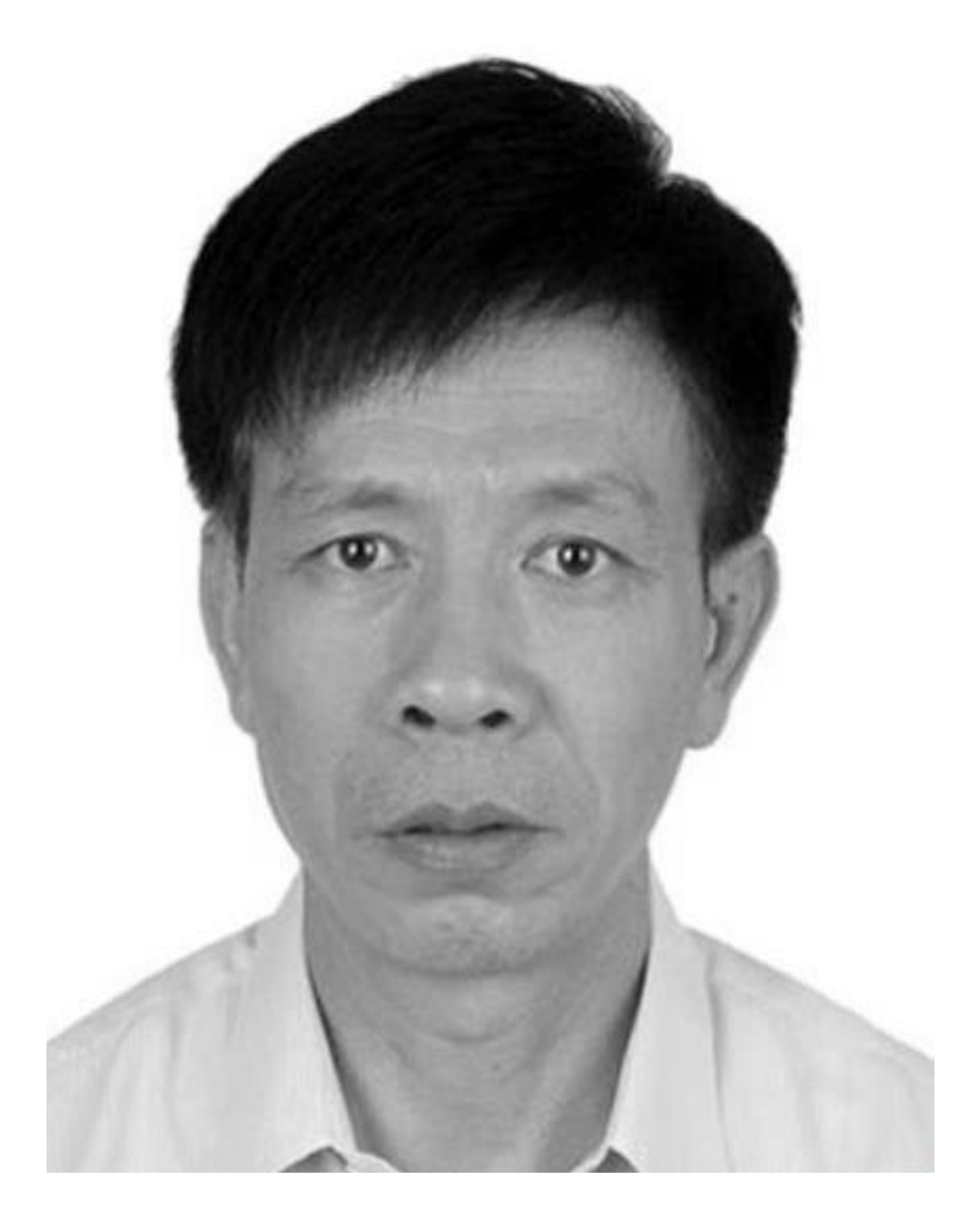}}]{Weihai Chen}(M'02)
	received the Ph.D. degrees from Beihang University, Beijing, China, 1996.
	He has been with the School of Automation Science and Electrical Engineering, Beihang University, as a Professor since 2007. He has authored or coauthored more than 200 technical papers in referred journals and conference proceedings and filed 18 patents. His research interests include bio-inspired robotics, computer vision, image processing, precision mechanism, automation, and control.
\end{IEEEbiography}
\vspace{-40pt}
\begin{IEEEbiography}[{\includegraphics[width=1in,height=1.25in,clip,keepaspectratio]{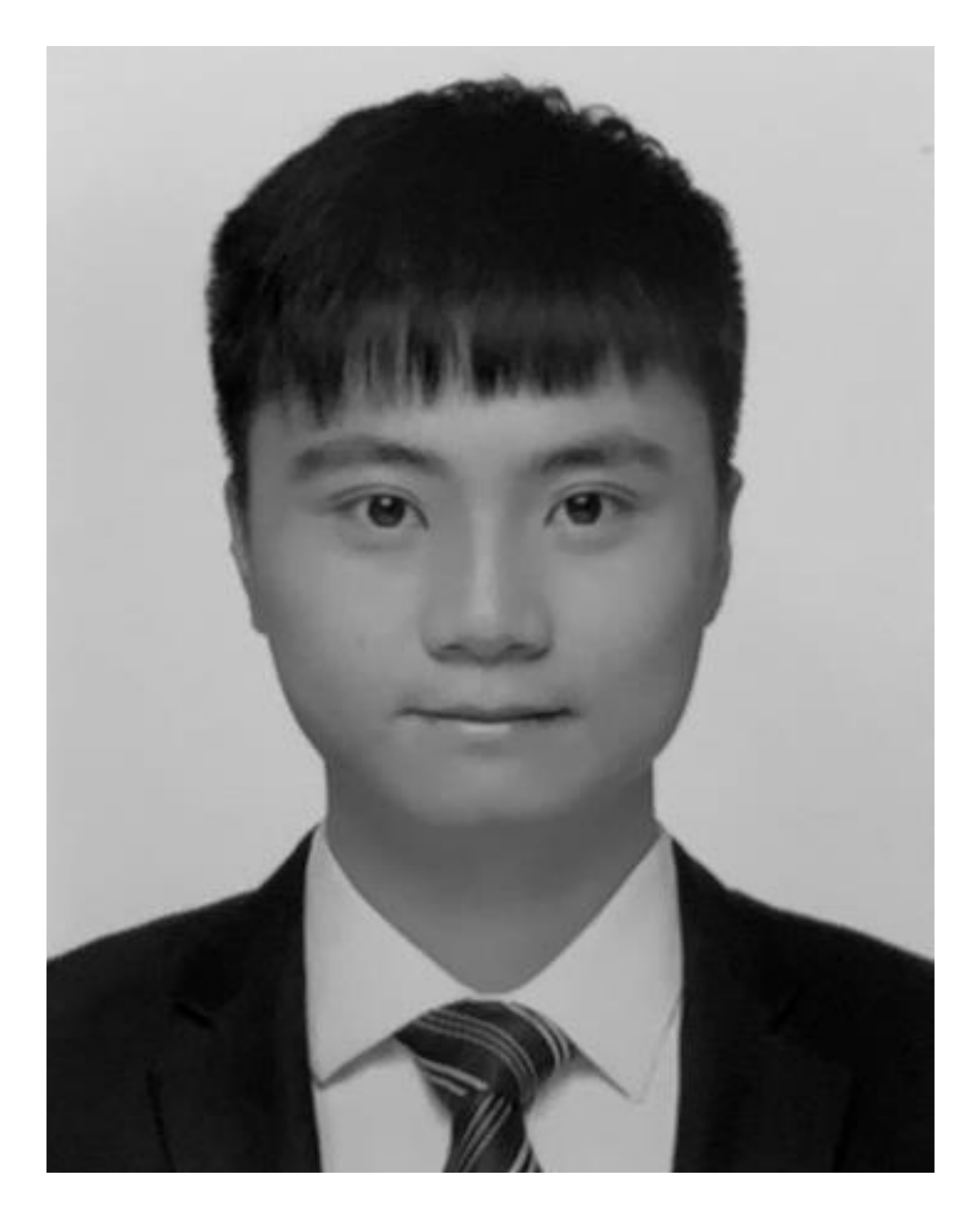}}]{Wei Wang}
	received the B.Sci. degree from North China Electric Power University, Beijing, China, in 2011, and the M.Eng. degree from Carnegie Mellon University, Pittsburgh, PA, USA, in 2013.
	He is currently a DPhil (PhD) student in computer science at the University of Oxford. His current research interests include robotics, machine learning and distributed system. 
\end{IEEEbiography}
\vspace{-40pt}
\begin{IEEEbiography}[{\includegraphics[width=1in,height=1.25in,clip,keepaspectratio]{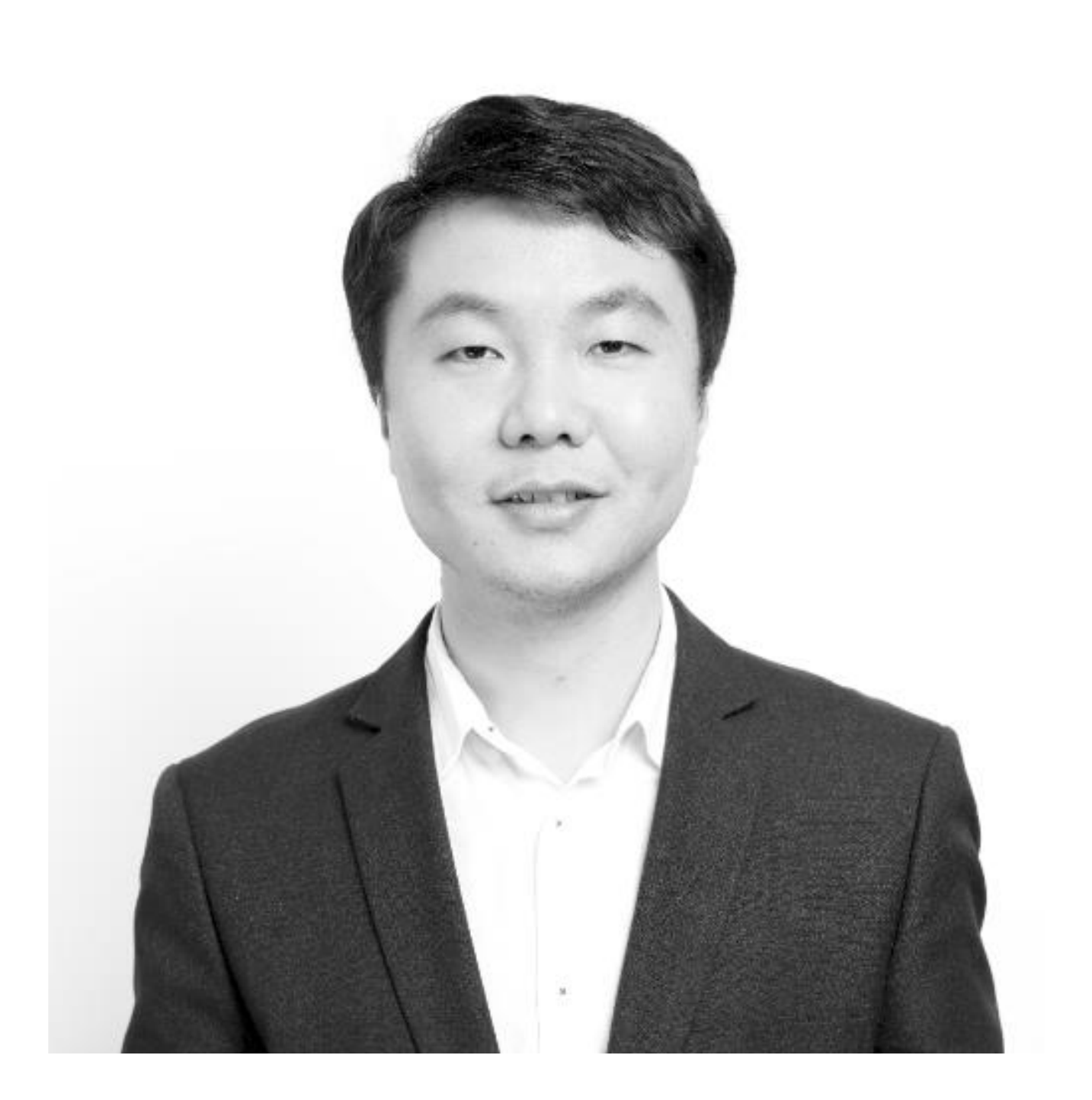}}]{Fanghong Guo}(M'16)
	received Ph.D. degree in Sustainable Earth from Energy Research Institute @NTU, Interdisciplinary Graduate School, Nanyang Technological University, Singapore in November 2016. 	
	He is currently with the Department of Automation, Zhejiang University of Technology, Hangzhou, China. His research interests include distributed cooperative control, distributed optimization on microgrid systems, and smart grid. 
\end{IEEEbiography}
\vspace{-40pt}
\begin{IEEEbiography}[{\includegraphics[width=1in,height=1.25in,clip,keepaspectratio]{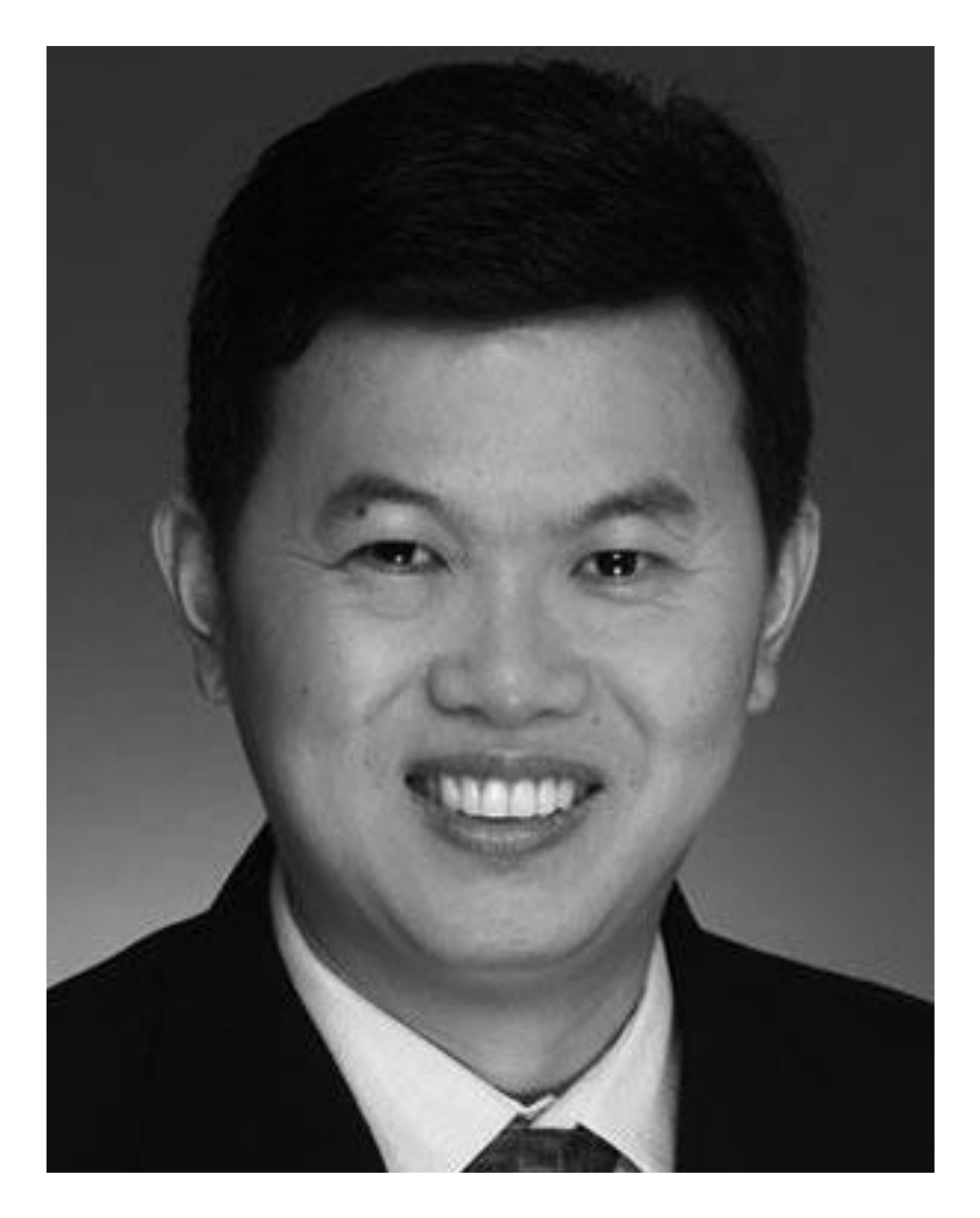}}]{Zhengguo Li}(A'01-M'03-SM'04) 
	received the Ph.D. degree from Nanyang Technological University, Nanyang, Singapore, in 2001. 
	His current research interests include mobile computational	photography, video processing and delivery, and switched and impulsive control systems. He is with the Agency for Science, Technology and Research, Singapore. He is an Associated Editor of the IEEE TRANSACTIONS ON IMAGE PROCESSING and served as an Associated Editor of the IEEE SIGNAL PROCESSING LETTERS. He served as a Technical Brief Co-Chair of SIGGRAPH Asia in 2012, the Workshop Chair of IEEE ICME 2013, an area Chair of IEEE ICIP 2016.
\end{IEEEbiography}


\vfill


\end{document}